\documentclass[letterpaper, 10 pt, conference]{ieeeconf}  

\IEEEoverridecommandlockouts                              

\usepackage{epstopdf}
\usepackage{times}
\usepackage{epsfig}
\usepackage{graphicx}
\usepackage{amsmath}
\usepackage{amssymb}
\usepackage{bm}
\usepackage[table,xcdraw]{xcolor}
\usepackage{booktabs}
\usepackage{multirow}

\usepackage{cite}
\makeatletter
\let\NAT@parse\undefined
\makeatother
\usepackage[pagebackref=true,breaklinks=true,letterpaper=true,colorlinks,bookmarks=false]{hyperref}

\title{\LARGE \bf
Learning 6-DoF Object Poses to Grasp  Category-level Objects by Language Instructions}

\author{Chilam Cheang$^{1}$, Haitao Lin$^{1}$, Yanwei Fu$^{1}$ and Xiangyang Xue$^{1}$
\thanks{*This work was supported in part by NSFC under Grant (No. 62076067), STCSM Project (19511120700), and Shanghai Municipal Science and Technology Major Project (No.2021SHZDZX0103).}%
\thanks{$^{1}$Fudan University. \{ccheang19,htlin19,yanweifu,xyxue\}@fudan.edu.cn. Yanwei Fu is the corresponding author, School of Data Science.}%
}

\begin{document}
\maketitle
\thispagestyle{empty}
\pagestyle{empty}

\def\eg{\textit{e.g.}}
\def\etal{\textit{et al.}}
\def\etc{\textit{etc.}}
\def\ie{\textit{i.e.}}

\newcommand{\revise}[1]{\textcolor[rgb]{0,0,0}{#1}}
\begin{abstract}
This paper studies the task of any objects grasping from the known categories by free-form language instructions. This task demands the technique in computer vision, natural language processing, and robotics. We bring these disciplines together on this open challenge, which is essential to human-robot interaction. Critically, the key challenge lies in inferring the category of objects from linguistic instructions and accurately estimating the 6-DoF information of unseen objects from the known classes. In contrast, previous works focus on inferring the pose of object candidates at the instance level. This significantly limits its applications in real-world scenarios.In this paper, we propose a language-guided 6-DoF category-level object localization model to achieve robotic grasping by comprehending human intention. To this end, we propose a novel two-stage method. Particularly, the first stage grounds the target in the RGB image through language description of names, attributes, and spatial relations of objects. The second stage extracts and segments point clouds from the cropped depth image and estimates the full 6-DoF object pose at category-level. Under such a manner, our approach can locate the specific object by following human instructions, and estimate the full 6-DoF pose of a category-known but unseen instance which is not utilized for training the model. Extensive experimental results show that our method is competitive with the state-of-the-art language-conditioned grasp method. Importantly, we deploy our approach on a physical robot to validate the usability of our framework in real-world applications. Please refer to the supplementary for the demo videos of our robot experiments~\footnote{Project webpage. \url{https://baboon527.github.io/lang_6d}}.
\end{abstract}


\vspace{-0.1in}
\section{Introduction}
    Understating natural language instruction is an essential skill for domestic robots, releasing humans from pre-defining a specific target for robot grasping by programming. This inspires the task of making robots understand human instructions. In this task, the robot demands to localize the target object by parsing the names, potential attributes, and spatial relations of objects from the language description. Thus it is non-trivial to make robotic grasping by linguistic description, as this task requires mature techniques from Computer Vision (CV), Natural Language Processing (NLP), and robotics. 
    In this paper, we bring these disciplines together on this open challenge, which is essential to human-robot interaction. 

    \begin{figure}[htbp]
     \centering
    \includegraphics[width=0.9\linewidth]{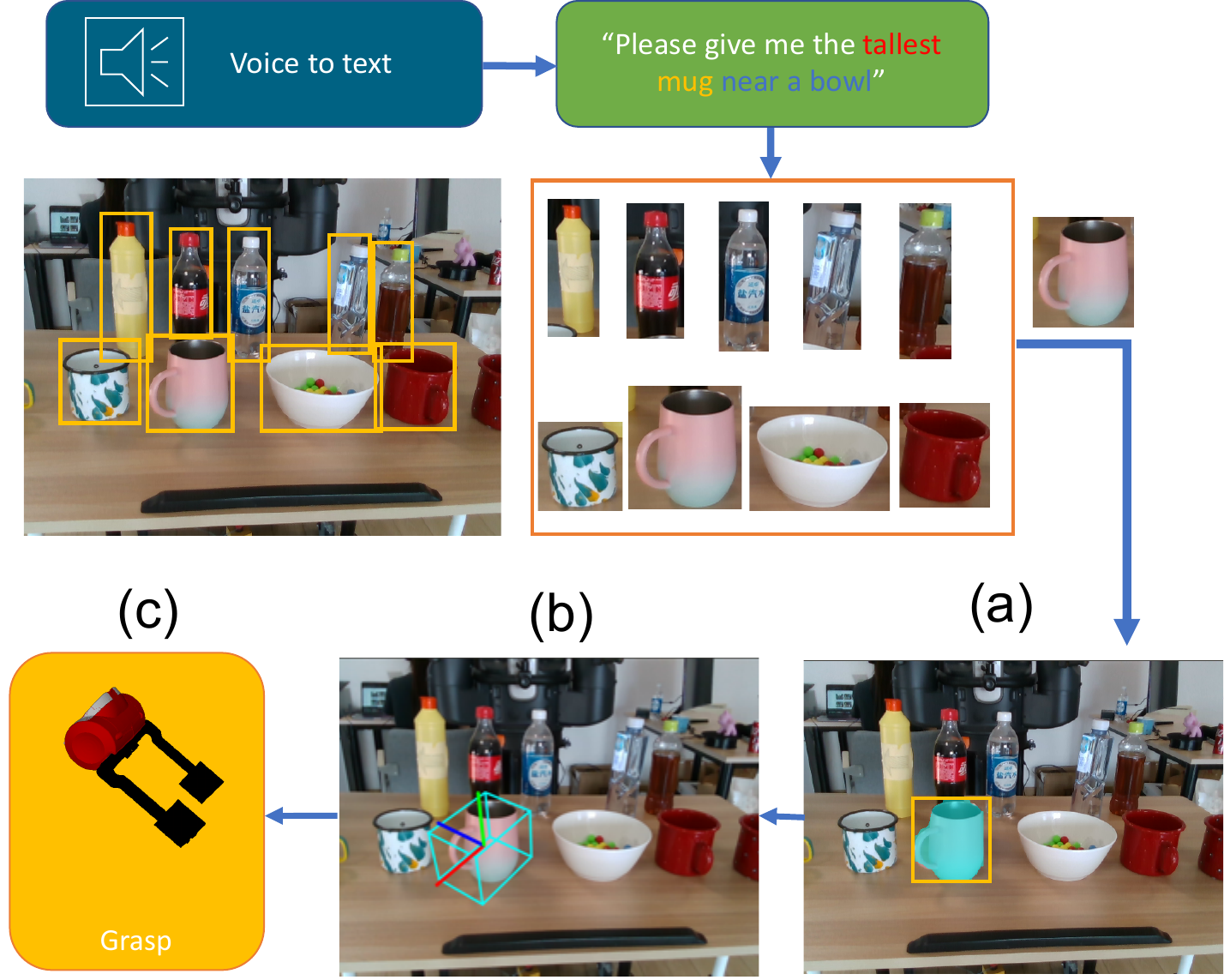}
    \vspace{-0.1in}
     \caption{\textbf{Task description of our framework.} Given an RGB-D image and a language description, we first (a) ground target 2D position in the image plane, and (b) infer the 6-DoF pose of the target object. Finally, (c) robotic grasping is implemented by utilizing the estimated 6-DoF object pose.
     \label{fig:task}}
    \vspace{-0.15in}
    \end{figure}

    The key challenge lies in inferring the category of objects from linguistic instructions, and accurately estimating the 6-DoF information of unseen objects from the known classes. Specifically, the vanilla object pose estimation approaches~\cite{xiang2017posecnn, 2012linemod, 2019densefusion} attempt to estimate the poses of objects from the image, while we aim at locating specific objects using a natural language description. Here we employ the  Bidirectional RNN from NLP to parse the linguistic instructions. 
    Furthermore, we focus on estimating the 6-DoF pose of the object at \textit{category-level}. That is to grasp any objects from the known categories, even though these objects have not been explicitly utilized for training the pose estimation models. We adopt the 6-DoF object pose representation~\cite{xiang2017posecnn, deng2020self}, which provides more abundant information to robots than representations of 2D  (oriented) bounding box. Especially,
     the 2D object representations are widely used by previous language-conditioned grasping methods~\cite{shridhar2018interactive, nguyen2020robot, guadarrama2014open} 
      to identify the objects at \textit{object-level}. This unfortunately greatly restricts the flexible choices of the optimal grasp poses for robots in complex scenes.
     
    
    
 

    
    
    \begin{figure*}[htbp]
     \centering
    \includegraphics[width=0.9\linewidth]{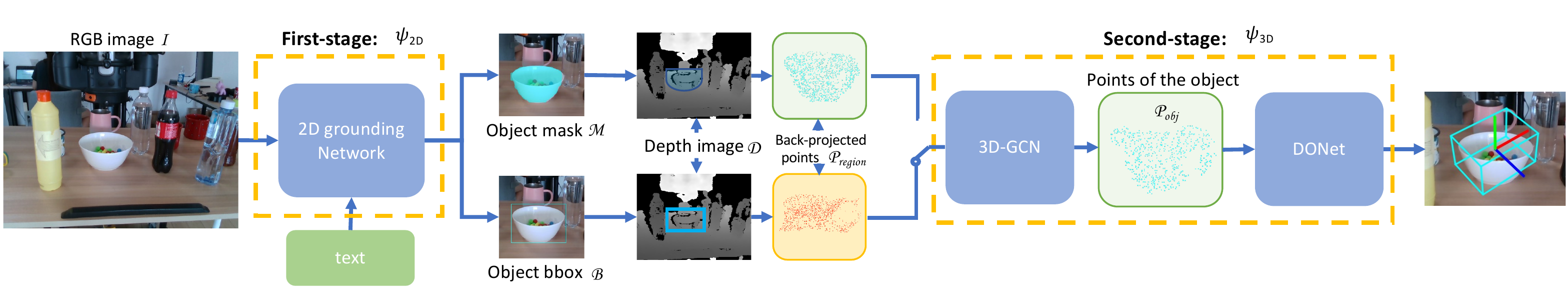}
    \vspace{-0.1in}
     \caption{\textbf{An overall pipeline of our model.} Given an RGB-D image and a description text, we first ground the 2D position in the image by a 2D grounding network. The grounding result is represented as a bounding box or a mask for cropping and back-projecting corresponding depth pixels into a 3D point cloud. 3D-GCN network further segments such a point cloud to get a more accurate object point cloud. Finally, object point cloud($\mathcal{P}_{obj}$) are fed into a category-level 6-DoF pose estimator - DONet for inferring the 6-DoF pose of the target object.}
     \label{fig:pipeline}
    \vspace{-0.15in}
    \end{figure*}

    In this paper, we propose a category-level 3D object localization model to grasp unseen instances via natural language description, as shown in Fig.~\ref{fig:task}. Formally, we present a two-stage approach to tackle this task. Take an RGB-D image and a natural language description as input, our goal is to infer the 6-DoF pose and size of the most likely object that matches the description. The first stage is to locate the 2D bounding box and generate the corresponding mask from the natural language description. At the second stage, cropped and back-projected region points are fed into the point cloud segmentation network, then the segment point cloud is used to estimate object pose and size simultaneously. In contrast to other 3D visual grounding approaches~\cite{chen2019scanrefer,achlioptas2020referit3d}, we estimate a 6-DoF object pose instead of solely predicting an 3D bounding box with the single-axis rotation. The 6-DoF pose representation provides more comprehensive information of 3D rotation, while the 3D bounding box only represents the rotation in a horizontal plane.

    To summarize, our paper makes several contributions: (1) We propose a systematic framework for category-level 3D visual grounding, and it can localize category-known but unseen instances which are not used for model training in the real-world 3D scene. (2) Our system enables matching most 2D detection and segmentation models to achieve 3D visual grounding or multi-object pose and size estimation. (3) We exploit the segmentation module to improve the quality of noise depth input. Experiments show that such a module gains better estimation results and performs more precise pose and size estimation from raw point cloud captured by a single-view camera. (4) We deploy our system in a physical Baxter Robot. The robotic experiment shows that the effectiveness of our system in accurate 3D localization for unseen instances in real-world scenarios.


\section{Related Work}
\label{sec:citations}
\noindent \textbf{2D visual grounding.} It targets localizing a specific object in the form of a 2D bounding box from an image guided by language description. Two-stage methods first generate 2D bounding box proposals by a 2D detector~\cite{redmon2018yolov3,ren2015faster}, then proposals are scored by the similarity of the given description. The proposal with the highest confidence score will be taken as the final localization of the target object.
    Although 2D visual grounding approaches~\cite{liu2019learning,ye2019cross,yang2020graph,liu2020learning,liu2020graph} precisely localize objects in the image plane, they fail to localize objects in the 3D scene. 
    In contrast, our method first ground objects in 2D for reducing computation cost; instead of diametrically localizing an object in 3D representation, we estimate the 6-DoF pose from masked points of the target.
    
\noindent \textbf{3D visual grounding.}
   It extends 2D visual grounding task to 3D space. Recent researches~\cite{chen2019scanrefer, liu2021refer,achlioptas2020referit3d,huang2021text,feng2021free} aim to localize objects in a 3D indoor scenario, which heavily relies on 3D proposals by utilizing PointNet++~\cite{qi2017pointnet++} or VoteNet~\cite{qi2019deep} as the backbone with increased computational cost. Furthermore, these methods identify objects in an indoor scenario by a 3D bounding box representation, containing 3D location, 3D size and one-axis rotation angle, This 3D bounding box representation has less degree-of-freedom to handle complex manipulation tasks since the orientation is restricted in the horizontal plane.
    Instead, 6-DoF pose representation can fully infer object in 6-DoF. This is amenable for most robot manipulation tasks than the previous works.
    
\noindent \textbf{6-DoF object pose estimation in robotic grasping.}
    Compared to 2D object localization~\cite{lenz2015deep,zhang2019roi}, 6-DoF pose provides more comprehensive information of target object in 3D space, facilitating  better robot manipulation~\cite{wada2020morefusion}. In terms of generalization ability, pose estimation methods can be roughly characterized as the groups of \textit{instance-level} and \textit{category-level}. 
    The former ones~\cite{2012linemod, 2016textureless, xiang2017posecnn,peng2019pvnet, 2019densefusion} rely on pre-scanned CAD models of objects, which is impractical in real-world scenarios.
    In contrast, the latter ones alleviate this reliance: 
    In \cite{wang2019normalized} they address category-level pose estimation problems and provide a large-scale corresponding benchmark. 
    A series of follow-up works~\cite{chen2020learning, tian2020shape, chen2021fs, lin2021donet} further improve the accuracy of this category-level task. 
    However, existing 6-DoF object pose estimation methods are still infeasible to distinguish the user-specified objects. In contrast, our method integrates the language with image features to localize a specific instance. Thus it can directly enable robot grasping by description from users.
    
\noindent \textbf{Language-conditioned Grasping.} The development of CV and NLP enables the robot to understand language commands. Previous works~\cite{guadarrama2014open, hatori2018interactively, yang2021attribute, shridhar2018interactive, nguyen2020robot} localize the target objects in the form of a 2D bounding box at object-level, which is conditioned on the class names, attributes, and spatial relations~\cite{guadarrama2014open, hatori2018interactively, yang2021attribute, shridhar2018interactive} or usage of description~\cite{nguyen2020robot}. Chen et al.~\cite{chen2021joint} propose a joint network that outputs satisfied 2D planar grasps from an RGB image at grasp-level. These works localize objects in a 2D planar plane by rough box representation. Instead, our method localizes objects in a 2D plane by pixel-level precise mask and predicts the full 6-DoF poses of the objects for better grasping.


\section{Method}
\label{sec:problem statement}
\subsection{Problem Statement}
Given an RGB-D image ($\mathcal{I}$, $\mathcal{D}$) and a natural language instruction $\mathcal{W}$, where $\mathcal{I} \in \mathbb{R}^{H \times W \times 3}$ is the RGB image and $\mathcal{D} \in \mathbb{R}^{H \times W \times 1}$ is the depth image. Our task is to estimate the 6-DoF pose of the `most likely' object from the language instruction $\mathcal{W}$. The 6-DoF object pose representation consists of translation vector $\bm{t} \in \mathbb{R}^3$ and rotation matrix $\bm{R}\in SO(3)$. 

Our framework is illustrated as in Fig.~\ref{fig:pipeline}. The RGB image $\mathcal{I}$ and language instruction $\mathcal{W}$ will be fed forward to a 2D object localization network at the first stage. For 2D object localization, several proposals will be generated by a region proposal network to localize potential candidates. Then the image, language description, and proposal bounding box info will be fed forward to a 2D grounding network to localize the best match proposal. The proposal with the highest confidence score will be taken as the  2D bounding box $\mathcal{B}$ of the target with a predicted category label $\textit{c}$. Then the bounding box $\mathcal{B}$ will be fed into the mask segmentation branch for predicting a pixel-wise segmentation mask $\mathcal{M}$. 
Totally, \textit{the first stage} (Stage 1) of language-guided 2D localization is denoted as:
\begin{equation}
    {\Psi}_{2D}:(\mathcal{I}, \mathcal{W})\rightarrow (\mathcal{B}/\mathcal{M}, c)
\end{equation}

Obtained the categorical 2D bounding box $\mathcal{B}$ or mask $\mathcal{M}$, we crop the corresponding pixels from the depth image $\mathcal{D}$, and the cropped depth pixels are further back-projected into 3D scene point cloud $\mathcal{P}_{region}$  with known camera intrinsic parameters. 
Then the filtered points $\mathcal{P}_{obj}$ of the target object, deriving from 3D segmentation network, are fed into the pose estimator for recovering the final 6-DoF object pose $(\bm{R}, \bm{t})$ and size $\bm{s}$. Therefore we have \textit{the second stage} (Stage 2)  as:
\begin{equation}
    {\Psi}_{3D}:(\mathcal{D}, \mathcal{B} / \mathcal{M}, c)\rightarrow (\bm{R},\bm{t};\bm{s})
\end{equation}

\begin{figure}
  \centering
\includegraphics[width=0.9\linewidth]{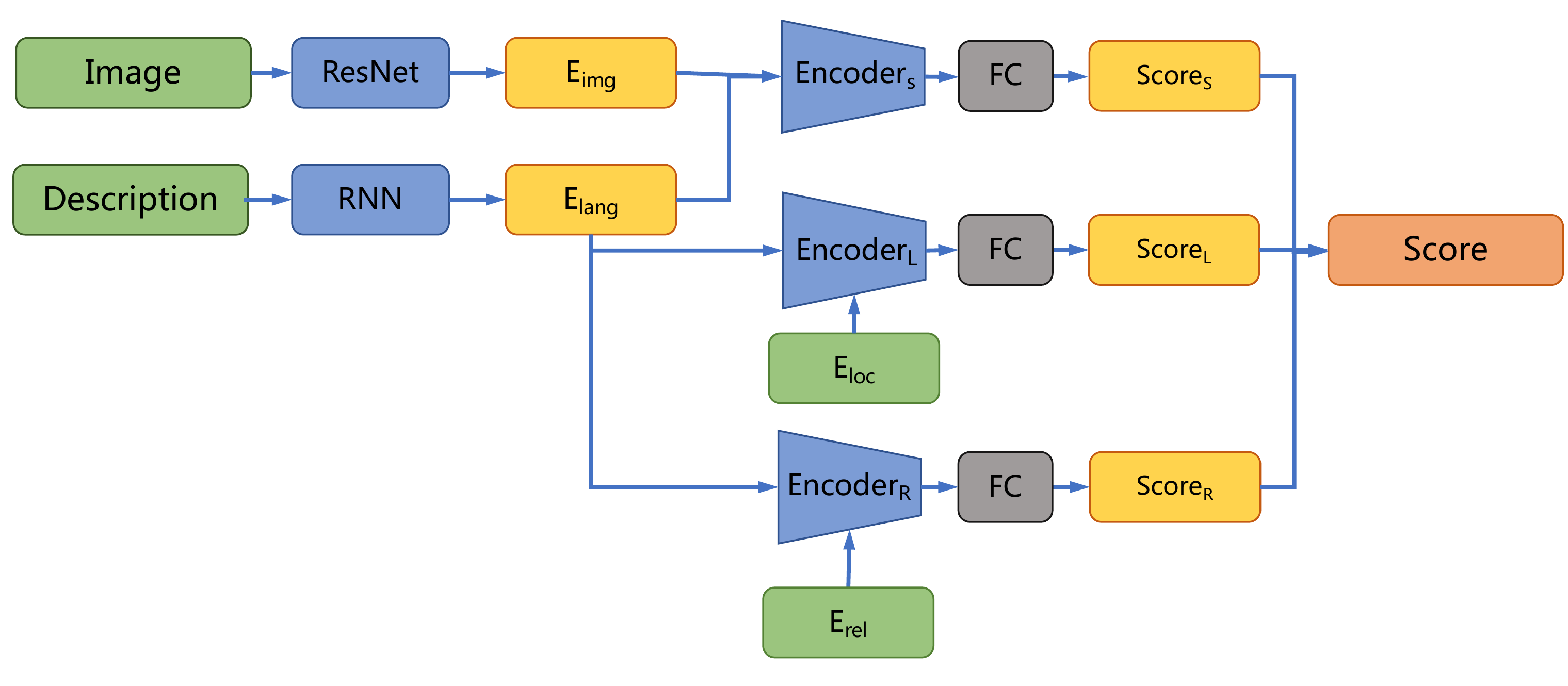}
\vspace{-0.1in}
  \caption{\textbf{Language Network of 2D Visual Grounding.} Language and visual features are extracted by RNN and ResNet, respectively. Features are then fused to predict the confidence score of the subject in the language description. Position annotation of bounding box is encoded as $E_{loc}$ for matching location information of bounding box and description. Neighbor bounding box location information is encoded as $E_{rel}$ for the prediction of relation scores.\label{language_network}} 
\vspace{-0.15in}
\end{figure}

\subsection{2D localization from natural language description}
To localize the 2D object via language instruction $\mathcal{W}$, object proposals are first generated to locate candidate target in \textit{visual network}. Then the language and visual features are utilized to infer the most matching object by \textit{language network}. For the \textit{visual network}, the image $\mathcal{I}$ is first fed onto a region proposal network (RPN) to predict potential region of interests in the form of bounding box $\mathcal{B}$. Corresponding object classes $\textit{c}$ and object masks $\mathcal{M}$ are then estimated based on a Mask-RCNN~\cite{he2017mask}. Inspired by~\cite{yu2018mattnet}, we follow the structure to estimate the confidence score of each proposal bounding box. Each proposal will be calculated with an overall score by three modules of the subject, location, and relation from \textit{language network}, as shown in Figure~\ref{language_network}. \revise{This exploits the Bidirectional RNN to learn weights of modules and language features. Besides, ResNet is used as a visual feature extractor; language feature $E_{lang}$ and visual feature $E_{img}$ are then fused and fed into the subject encoder $Encoder_{S}$ to estimate the confidence score of the referring subject($Score_{S}$). $E_{lang}$ and location annotation $E_{loc}$ are fused and fed into Location encoder $Encoder_{L}$ to estimate the confidence score of the location information in language description($Score_{L}$). $E_{lang}$ and Neighbor location information $E_{rel}$ are fed into encoder $Encoder_{R}$ to estimate the confidence score of target relationship($Score_{R}$). The weighted sum is calculated for candidates, and the best-matched mask with the highest confidence score will be taken as the final 2D localization of the target object. We train the language-conditioned 2D image detector on the RefCOCO dataset, as this dataset provides diverse textual descriptions and images from MSCOCO dataset. Additionally, we adopt the combined hinge loss for the 2D grounding network as in~\cite{yu2018mattnet}.}

\subsection{3D segmentation}
\label{sec:3d seg}
After obtaining the 2D localization of the target object, we further extract the target point cloud $\mathcal{P}_{obj}$ of the object from the depth image to estimate pose. 
Particularly, the predicted 2D bounding box $\mathcal{B}$ or mask $\mathcal{M}$ from the first stage is employed to crop the corresponding pixels from depth image $\mathcal{D}$. The cropped pixels are then back-projected into region points $\mathcal{P}_{region}$ by using the known intrinsic camera parameters. However, the region points $\mathcal{P}_{region}$ usually contain the foreground points of the object and outliers of background points. 
Such outliers may influence the performance of the 6-DoF pose estimation network. Therefore, we utilize the off-the-shelf 3D segmentation network 3D-GCN~\cite{lin2020convolution} as it is designed for object part segmentation. The 3D-GCN prepossess the back-projected points $\mathcal{P}_{region}$ for outliers removal, and gets the final points of the object $\mathcal{P}_{obj}$.

Concretely, we leverage the NOCS dataset~\cite{wang2019normalized} to generate our synthetic data for training the segmentation network. We design two specific strategies for synthetic data generation to efficiently train our network to differentiate the foreground and background points. (1) Given a 2D bounding box of the target object, the cropped points usually contains object and background part.
For the first strategy, we randomly expand the 2D ground-truth bounding box of the object to simulate the situation, manually introducing different proportions of outliers from the background. (2) For the second strategy, the accurate pixel-wise mask of the target objects is given; most of the noisy points are introduced from the silhouette part of the masks. To train the network robust to such a situation, we dilate the ground-truth object mask to generate a handful of the outliers of background points around the object.

Further, due to the domain gap between the real-world point clouds and the synthetic ones, we add Gaussian noise $\mathcal{N}(\bm{\mu}, \bm{\sigma}^2)$ to the synthetic back-projected points for bridging the reality gap; $\bm{\mu}=\bm{0}$, ${\sigma}_x = {\sigma}_y=0.001$ , and ${\sigma}_z =0.001$, where $\bm{\sigma}=({\sigma}_x,{\sigma}_y,{\sigma}_z)$. This is validated when transferring the model to real-world scenarios, as shown in ablation study in Sec.~\ref{sec:ablation study}.

\subsection{Category-level pose estimation}
For 6-DoF poses of category-level objects, we follow the definition of category-level object pose as illustrated in ~\cite{wang2019normalized}. 
Concretely, we estimate the 6-DoF object pose by utilizing the DONet~\cite{lin2021donet} pose estimator, and then execute grasp according to the predicted poses of the target instances. 
Category-level pose estimator takes the predicted category and point cloud of the object as input, and outputs the 6-DoF object pose and 3D size in the camera frame. The pose estimator is only trained on synthetic data but capable of effectively transferring to real-world applications. DONet pose estimator is trained on the single object points which are back-projected from the rendered depth map of CAD models in the ShapeNet repository. As DONet is a point-based method that only takes the point cloud of the object as input, the noisy points will slightly bias the inference results. Correspondingly, integrated with the 3D segmentation network, our system can better perform real-world tasks.

\section{Experiment}
\label{sec:experiment}
We design three-part experiments to evaluate the efficacy of our framework, including 2D natural language grounding, ablation study on our key components, and real robot grasping by language instructions.

\noindent \textbf{Dataset}. RefCOCO\cite{kazemzadeh2014referitgame} benchmark is used for evaluation of 2D visual grounding accuracy. The grounding accuracy of validation, testA and testB sets are evaluated to measure the performance. To validate the accuracy of pose estimation of our framework, we use the widely-used benchmark NOCS dataset~\cite{wang2019normalized}. The dataset cover six categories-\textit{bottle}, \textit{bowl}, \textit{camera}, \textit{can}, \textit{laptop} and \textit{mug}. We use the NOCS-REAL275 which has 8K welled-annotated real-world data. Different from the setting as \cite{wang2019normalized, tian2020shape}, we only utilize the synthetic data for training our network. For real-world robotic experiments, we choose 33 household objects covers 3 categories including \textit{bottle}, \textit{bowl}, and \textit{mug}.

\noindent \textbf{Evaluation metrics.} {(1) 2D natural language grounding.} For the evaluation of 2D grounding, accuracy is used to measure the performance of object localization. If the localization result is consistent with ground-truth annotation, the result will be evaluated as a true positive sample. The accuracy of language-guided grounding indicates the successful localization cases in testing image sets. 
{(2) 6-DoF object pose and size estimation.} For quantitative comparison of the pose estimation, we adopt the metric as~\cite{wang2019normalized, tian2020shape, chen2021fs}. Typically, we report the intersection over union (IoU) metric under different threshold $k\%$, noted as $IoU_k$ for 3D object detection. As for pose recovery, we evaluate the 6-DoF pose estimation errors that is less than $n^{\circ}$ for 3D rotation and $m$ cm for 3D translation, denoted as $n^{\circ} m $cm.

\begin{table}[t]
\centering
\footnotesize
\caption{Results of 2D Grounding accuracy for different methods in the RefCOCO Dataset. \label{tab:nlp_eval} } 
\vspace{-0.1in}
\renewcommand\tabcolsep{3pt}
\renewcommand{\arraystretch}{0.85}
\begin{tabular}{c|ccc}
\toprule[1pt]
{} & UMD Refexp~\cite{nagaraja2016modeling} & INGRESS~\cite{shridhar2020ingress} & Ours \\
\midrule[0.5pt]
val & 75.5 & 77.0 & \textbf{85.32} \\
testA & 74.1 & 76.7 & \textbf{85.91} \\
testB & 76.8 & 77.7 & \textbf{83.87} \\
\bottomrule[1pt]
\end{tabular}
\vspace{-0.15in}
\end{table}

\begin{table}[]
\caption{Comparisons of variants of our systems to validate the importance of 3D segmentation module. 
\label{tab:ablation segmentation}}
\vspace{-0.1in}
\footnotesize
\centering
\renewcommand\tabcolsep{1pt}
\renewcommand{\arraystretch}{0.85}
\begin{tabular}{cccc|c|ccccc}
\toprule[1pt]
\multicolumn{5}{c|}{Method} &
  \multirow{4}{*}{$IoU_{50}$} &
  \multirow{4}{*}{$IoU_{75}$} &
  \multirow{4}{*}{$5^{\circ}2cm$} &
  \multirow{4}{*}{$5^{\circ}5cm$} &
  \multirow{4}{*}{$10^{\circ}2cm$}  \\ \cline{1-5}
\multicolumn{4}{c|}{2D localization} & \multirow{3}{*}{Seg.} &  &  &  &  &    \\ \cline{1-4}
\multicolumn{2}{c|}{bbox}            & \multicolumn{2}{c|}{mask} &  &  &  &  &  &  \\ \cline{1-4}
{yolo} & Ours(B) & {mrcnn} &  Ours(M) & &  &  &  &  &   \\ \midrule[0.5pt]
\checkmark &  &  &  & $\times$ & {46.2}  & {19.3} & 6.2  & 14.6 & 13.1  \\
\checkmark &  &  &  & \checkmark & \textbf{67.1}  & \textbf{48.3} & \textbf{33.3}  & \textbf{42.2} & \textbf{52.5}  \\ \midrule[0.5pt]
& \checkmark &  &  & $\times$  & {40.9}  & {16.5} & 5.3  & 10.8 & 11.7  \\
& \checkmark &  &  & \checkmark & \textbf{62.3} & \textbf{43.5} & \textbf{22.5} & \textbf{28.2} & \textbf{40.1}  \\ \midrule[0.5pt]
&  & \checkmark &  & $\times$ & \textbf{82.8}  & {71.9}  & {32.9} & {42.8}& {56.0}\\
&  & \checkmark &  & \checkmark & 81.5 & \textbf{69.6} & \textbf{37.5} & \textbf{45.4} & \textbf{61.8} \\ \midrule[0.5pt]
&  &  & \checkmark & $\times$  & 68.9 & \textbf{55.6} & 23.2 & 30.8 & 42.3  \\
&  &  & \checkmark & \checkmark & \textbf{69.0} & {53.9} & \textbf{28.3} & \textbf{34.0} & \textbf{47.2}   \\ \bottomrule[1pt]
\end{tabular}
\vspace{-0.15in}
\end{table}

\begin{table}[t]
\centering
\renewcommand\tabcolsep{1pt}
\renewcommand{\arraystretch}{0.85}
\footnotesize
\caption{Comparisons of variants of our model by using different synthetic data generation strategies. \label{tab:ablation strategy}}
\vspace{-0.1in}
\begin{tabular}{c|ccc|cccccc}
\toprule[1pt]
{2D} & \multicolumn{3}{c|}{Strategies} & \multirow{2}{*}{$IoU_{50}$} & \multirow{2}{*}{$IoU_{75}$} & \multirow{2}{*}{$5^{\circ}2cm$} & \multirow{2}{*}{$5^{\circ}5cm$} & \multirow{2}{*}{$10^{\circ}2cm$} \\
Input & bbox & mask & noise &  &  & \\
\midrule[0.5pt]
\multirow{4}{*}{Bbox} & \checkmark & & & 25.6 & 6.0 & 1.2 & 1.8  & 3.9   \\
& & \checkmark & & 48.2  & 17.6  & 2.9   & 6.2 & 10.1  \\
& \checkmark & \checkmark & & 57.2  & 23.3  & 3.6 & 6.8 & 11.9  \\
& \checkmark & \checkmark & \checkmark & \textbf{62.3} & \textbf{43.5} & \textbf{22.5}
& \textbf{28.2} & \textbf{40.1} \\ \midrule[0.5pt]
\multirow{4}{*}{Mask} & \checkmark & & & 18.7  & 2.9 & 0.6 & 1.4  & 2.1 \\
& & \checkmark & & 38.6  & 12.5  & 2.5 & 5.4 & 8.2  \\
& \checkmark & \checkmark & & 57.2  & 21.1  & 3.9 & 8.1 & 10.9  \\
& \checkmark & \checkmark & \checkmark & \textbf{69.0} & \textbf{53.9} & \textbf{28.3} & \textbf{34.0} & \textbf{47.2}  \\
\bottomrule[1pt]
\end{tabular}
\vspace{-0.15in}

\end{table}

\begin{figure*}[t]
 \centering
\includegraphics[width=0.88\linewidth]{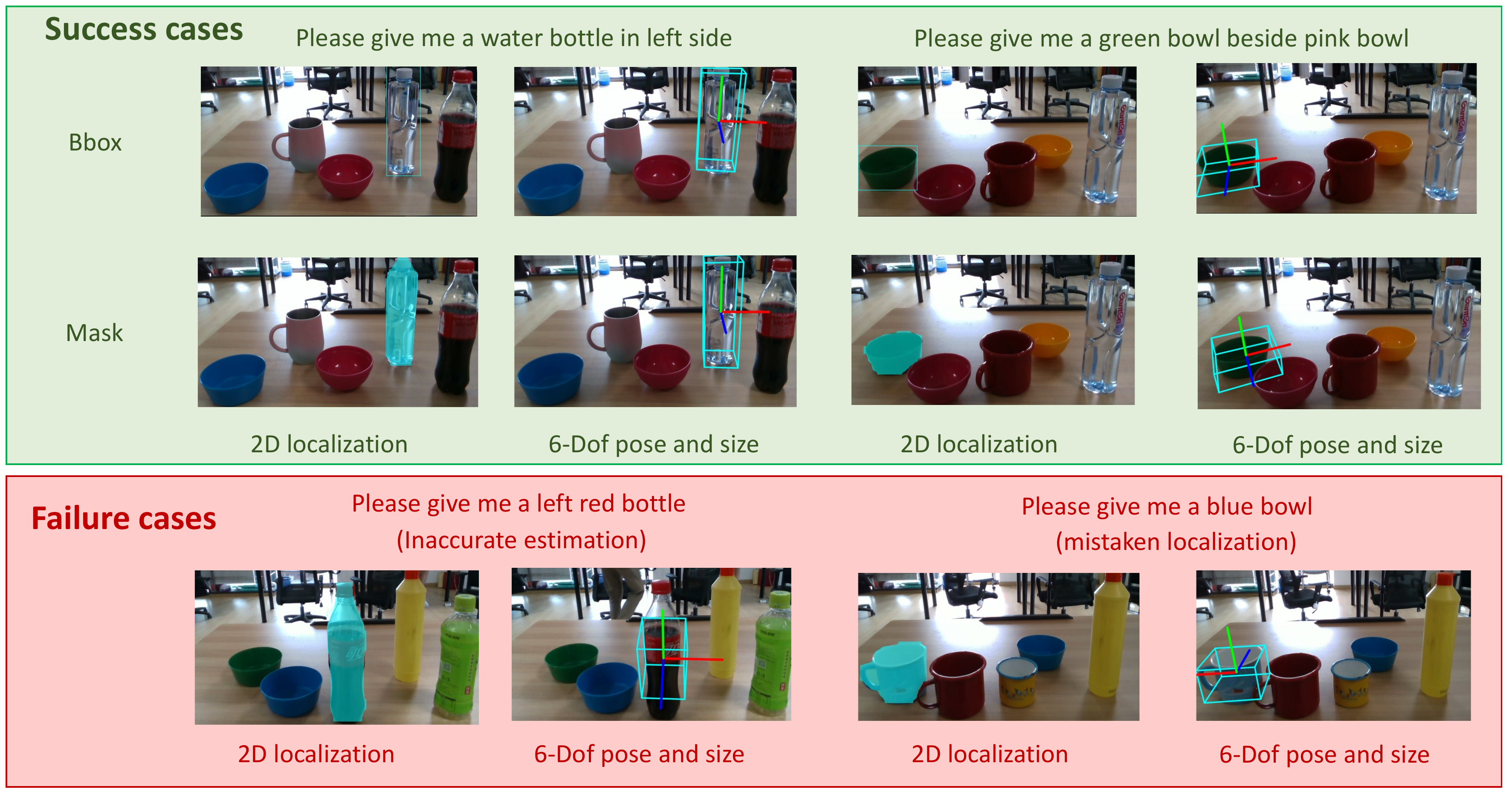}
 \vspace{-0.1in}
 \caption{\textbf{Visualization results from variants of our systems in real robot manipulation.} The top two rows are success cases of robotic grasp implementation by using language descriptions, in the form of the bounding box, and mask representations, respectively. The last two rows show the failure cases due to inaccurate pose estimation and mistaken localization.  \label{robot_eval1} }
 \vspace{-0.15in}
\end{figure*}

\subsection{2D Language Grounding Evaluation}

\noindent \textbf{Baseline methods.}
To validate the performance of our language grounding, we compare against different baseline methods. We compare the 2D grounding accuracy in RefCOCO\cite{kazemzadeh2014referitgame} with UMD Refexp~\cite{nagaraja2016modeling} and INGRESS\cite{shridhar2020ingress}.

\noindent \textbf{Results on the RefCOCO benchmark.} We compare the accuracy of UMD Refexp, INGRESS, and our language network as illustrated in Tab.~\ref{tab:nlp_eval}. In the evaluation of the 2D grounding network, the language network is tested with ground-truth proposals. We evaluate the model on the testing sets of Validation, testA, and testB, containing 1500, 750, and 750 images, respectively. It evident that our language grounding model has higher performance than other two baseline methods. 

\subsection{Ablations}
\label{sec:ablation study}
\noindent \textbf{Baseline methods.}
(1) We show \textit{variants of our systems} with using different 2D localization input, and with or without 3D segmentation network (Seg.) as in Tab~\ref{tab:ablation segmentation}. For different 2D localization input, we use bounding box inputs from YOLO~\cite{redmon2018yolov3} or Ours(B) of bounding box input, and segmentation masks from MRCNN~\cite{he2017mask} or Ours(M) of mask input. 
(2) In Tab.~\ref{tab:ablation strategy}, we show \textit{variants of our models} trained on the data processed by different cropping strategies, \ie, bounding box, mask, and with or without Gaussian noise.

\noindent \textbf{Importance of 3D segmentation.} We utilize the DONet~\cite{lin2021donet} as our category-level pose estimator, and DONet model is solely trained on synthetic data. In a real-world deployment, background outliers exist that slightly decrease the performance of the pose estimator. Thus we add a 3D segmentation module to enhance the ability of DONet in real-world applications. 
As shown in Tab.~\ref{tab:ablation segmentation}, we compare the results of the \textit{variants of our systems}, it is evident that the 3D segmentation module further improves the accuracy of the 6-DoF pose estimation. \revise{Moreover, benefit from the application of segmentation module, results show similar accuracy in both bounding box and mask representation, and enables our system matching most 2D detection and segmentation models to implement single or multi-object pose estimation.}
3D segmentation module is significant when the input is bounding box representations, as DONet can only handle the input of mask representations. The bounding box representations introduce a amount of outliers from the background, thus biases the final performance.
Although the 3D segmentation module slightly decreases the performance of $IoU_k$ under mask representations, it gains superior enhancement on 6-DoF pose estimation. Overall, the 3D segmentation improves the performance of the 6-DoF pose and 3D size estimation.  

\noindent \textbf{Different synthetic data generation.}
Different synthetic data generation strategies are designed to deal with different kinds of 2D inputs, \ie, 2D bounding box and 2D segmentation mask. We explore the performance of different cropping mechanisms on the segmentation network. As shown in Tab.~\ref{tab:ablation strategy}, simultaneously using two types of cropping strategies has better performance than only using a single one.
Moreover, we surprisingly find that adding Gaussian noise to the synthetic data enjoys the good generalization in real-world deployment as in Tab.~\ref{tab:ablation strategy} (bbox+mask+noise). The 3D segmentation network is trained on the synthetic data. The density of the synthetic point cloud is usually uniform; thus, the segmentation model has high accuracy on synthetic testing data. Due to the various density of captured real-world point clouds, the performance of the model decreases sharply on real-world data. 
Overall, the performance of the segmentation network directly influences final results, as the lousy segmentation may remove the foreground points. The lacked foreground points, \eg, points of mug handle, ruin the crucial details for better pose and size recovery.

\begin{table*}[t]
\centering
\footnotesize
\caption{Comparisons on language-guided robotic grasping by different methods. $L_{Auc}$ is the  of 3D localization accuracy and $G_{Suc}$ is the grasping success rate. \label{tab:robot_eval_tab}}
\vspace{-0.1in}
\renewcommand\tabcolsep{6pt}
\renewcommand{\arraystretch}{0.85}
\begin{tabular}{c|c|cc|cc|cc|cc|cc|cc}
\toprule[1pt]
\multirow{2}{*}{2D Input} & \multirow{2}{*}{Categories} & \multicolumn{2}{c|}{Scene1} & \multicolumn{2}{c|}{Scene2} & \multicolumn{2}{c|}{Scene3} & \multicolumn{2}{c|}{Scene4} & \multicolumn{2}{c|}{Scene5} & \multicolumn{2}{c}{} \\
{} & {} & $L_{Auc}$ & $G_{Suc}$ & $L_{Auc}$ & $G_{Suc}$ & $L_{Auc}$ & $G_{Suc}$ & $L_{Auc}$ & $G_{Suc}$ & $L_{Auc}$ & $G_{Suc}$ &  &  \\
\midrule[0.5pt]
\multirow{3}{*}{INGRESS~\cite{shridhar2020ingress}} & Bowl & 88.8 & 74.07 & 66.67 & 66.67 & 75 & 75 & 75 & 66.67 & 83.33 & 72.22 &  &  \\ 
{} & Mug & 75 & 66.67 & 75 & 58.33 & 83.33 & 77.78 & 66.67 & 59.26 & - & - &  &  \\
{} & Bottle & 100 & 83.33 & 66.67 & 55.55 & 66.67 & 66.67 & 75 & 75 & 55.55 & 55.55 &  & \\
\midrule[0.5pt]
\multirow{3}{*}{Ours(Bbox)} & Bowl & 100 & 81.22 & 100 & 100 & 50 & 50 & 25 & 25 & 83.33 & 83.33 &  &  \\ 
{} & Mug & 50 & 50 & 50 & 50 & 100 & 100 & 88.89 & 81.33 & - & - &  &  \\
{} & Bottle & 100 & 83 & 100 & 88.66 & 100 & 94.33 & 75 & 58 & 100 & 100 &  & \\
\midrule[0.5pt]
\multirow{3}{*}{Ours(Mask)} & Bowl & 100 & 85 & 100 & 100 & 50 & 50 & 25 & 25 & 83.33 & 77.67 &  &  \\ 
{} & Mug & 50 & 50 & 50 & 50 & 100 & 100 & 88.89 & 88.89 & - & - &  &  \\
{} & Bottle & 100 & 91.5 & 100 & 100 & 100 & 94.33 & 75 & 75 & 100 & 96.22 &  &  \\
\midrule[0.5pt]
\midrule[0.5pt]
\multirow{2}{*}{2D Input} & \multirow{2}{*}{Categories} & \multicolumn{2}{c|}{Scene6} & \multicolumn{2}{c|}{Scene7} & \multicolumn{2}{c|}{Scene8} & \multicolumn{2}{c|}{Scene9} & \multicolumn{2}{c|}{Scene10} & \multicolumn{2}{c}{Overall} \\
{} & {} & $L_{Auc}$ & $G_{Suc}$ & $L_{Auc}$ & $G_{Suc}$ & $L_{Auc}$ & $G_{Suc}$ & $L_{Auc}$ & $G_{Suc}$ & $L_{Auc}$ & $G_{Suc}$ & $L_{Auc}$ & $G_{Suc}$ \\
\midrule[0.5pt]
\multirow{3}{*}{INGRESS~\cite{shridhar2020ingress}} & Bowl & 100 & 100 & - & - & 100 & 100 & 75 & 50 & 77.78 & 77.78 & 82.00 & 75.33 \\ 
{} & Mug & 100 & 100 & 66.67 & 61.11 & 66.67 & 66.67 & 83.33 & 83.33 & 100 & 100 & 77.55 & 72.79 \\
{} & Bottle & 77.78 & 66.67 & 88.89 & 81.48 & 83.33 & 72.22 & 66.67 & 55.55 & 100 & 100 & 77.78 & 69.31\\
\midrule[0.5pt]
\multirow{3}{*}{Ours(Bbox)} & Bowl & 100 & 100 & - & - & 100 & 100 & 100 & 83.3 & 100 & 100 & {88.24} & {82.35} \\ 
{} & Mug & 75 & 66.7 & 83.3 & 83.3 & 83.3 & 83.3 & 100 & 100 & 75 & 66.67 & {81.63} & {79.59} \\
{} & Bottle & 100 & 74.07 & 100 & 96.30 & 100 & 100 & 66.67 & 66.67& 100 & 100 & {95.16} & {86.56} \\

\midrule[0.5pt]
\multirow{3}{*}{Ours(Mask)} & Bowl & 100 & 100 & - & - & 100 & 100 & 100 & 75 & 100 & 100 & {88.24} & {83.36} \\ 
{} & Mug & 75 & 66.7 & 83.3 & 83.3 & 83.3 & 83.3 & 100 & 100 & 75 & 75 & {81.63} & {80.27} \\
{} & Bottle & 100 & 85.19 & 100 & 96.30 & 100 & 83.3 & 66.67 & 61.11 & 100 & 100 & 95.16 & 90.86\\
\midrule[0.5pt]
\multicolumn{12}{c|}{Average Results of 10 Scenes (INGRESS)}  & 78.40 & 72.22\\
\midrule[0.5pt]
\multicolumn{12}{c|}{Average Results of 10 Scenes (Input of Bbox)}  & 88.34 & 82.83\\
\midrule[0.5pt]
\multicolumn{12}{c|}{Average Results of 10 Scenes (Input of Mask)}  & 88.34 & 84.83\\
\bottomrule[1pt]
\end{tabular}

\vspace{-0.15in}
\end{table*}

\subsection{Real robot evaluation}
Our method aims to interact with a human more generally by using a language description of the target object. Thus, we deploy our algorithms in the physical robot to validate the capability of the proposed framework. 

\subsubsection{Setup}
The Baxter robot has dual 7-DoF robot arms with parallel grippers mounted with a calibrated RealSense D415 camera on its base, and our model is deployed on a desktop with a single \textit{NVIDIA GTX 1070} GPU. We implement robotic grasping by dual robotic arms. The width of the gripper on the left arm ranges from 0cm to 2.5cm, and the right gripper from 2cm to 6cm. Different travel distances of the gripper are used to adapt different kinds of grasping.

\subsubsection{Configuration of the test scenes}
We collect some household objects and design experiments to localize specific instances by language descriptions to evaluate the localization accuracy. 
In our experiments, different bowls, mugs, and bottles (totaling 33 objects) with large intra-class variations in color, texture, and geometry shape are prepared for real robot experiment. Concretely, we design 10 scenes to tackle \textit{target-oriented robotic grasping task} for evaluation. Each scene contains 5 objects randomly selected from category of \textit{bottle}, \textit{bowl}, and \textit{mug} with large shape variations.
Each scene contains 5 objects randomly selected from category of \textit{bottle}, \textit{bowl}, and \textit{mug}. We randomly select and place five objects in the view of the RGB-D camera. 
Given natural language description, the robot grasps the object that is most likely to match the description.

 \subsubsection{Principle of language generation}
As for language description, each description consists of \textit{category of the instance}, \textit{attributes} and \textit{relation between object regions}. 
Descriptions are generated with the different prefix including `Please give me a', `Hand \underline{~~~~} to me', `Grasp \underline{~~~~}' `Pick  \underline{~~~~} to me', `Pass me  \underline{~~~~}', and  `Give me \underline{~~~~}'. We design such free-form descriptions as input to further validate the capability of our model in handling unconstrained language instructions.
Five types of description will generate: \revise{(a). the only category name is given if only one instance belongs to the category in a scene, (b). category name with attribute description (\eg, red, transparent), (c).category name with the absolute location description (\eg, rightmost, middle) (d). category name with relation description (\eg, behind mug, between bottles), and (e). category name with attribute and relation descriptions or absolute location description. 
Follow the above principle of language description generation, our system localizes the target object, and then recovers the pose and size of the object. }

\subsubsection{Results}
Figure~\ref{robot_eval1} illustrates the visualization results of variants of our systems tested in various real-world scenarios. Our system localizes the unseen instance from a natural language description and executes the robotic grasping according to the estimated 6-DoF object pose and size. 
Take either 2D bounding boxes or masks as input, and our system outputs accurate 6-DoF object pose and size as in success cases of Fig.~\ref{robot_eval1}. The 2D input of masks results in better performance than that of bounding boxes, as the mask input filters the negative results of pixels to remove more outliers from the background.

We report the quantitative results of our robot experiments in Tab.~\ref{tab:robot_eval_tab}. Localization accuracy ($L_{Auc}$) is define as $L_{Auc} = T_{Localize}/(T_{Localize} + F_{Localize})$. $T_{Localize}$ is the amount of positive result, and $F_{Localize}$ is the amount of negative result of localization. The success rate of robotic grasping($G_{suc}$) is define as $G_{suc} = S_{grasp}/(S_{grasp} + F_{grasp})$. $S_{grasp}$ is the sum of successful grasp cases, and $F_{grasp}$ is the sum of failure grasp cases. The success rate of robotic grasping is highly dependent on the accuracy of localization. Quantitative results show that our model can precisely estimate 6-DoF object pose and achieve a desirable grasping success rate compared to the method of INGRESS. Particularly, failure cases of grasping also result from collisions during grippers approaching the object, as objects are too close. Overall, the physical robot evaluation demonstrates that our system is robust in different real-world scenarios.

\noindent \textbf{Failure cases:} Imprecise object size estimation in failure case of Fig.~\ref{robot_eval1} is caused by severe defects of point cloud input of the transparent object. Object localization may fail if multiple objects share similar attributes in the same scene. Take the last case in Fig.~\ref{robot_eval1} as an example, and we want to target the blue bowl behind the yellow mug. If the description only contains attribute and category name of the object, without giving any information of location and relation with other objects, the weight of \textit{attribute} might be greater than that of \textit{category name} and lead to mistaken localization.


\section{Conclusion}
\label{sec:conclusion}
This paper presents a novel systematic framework capable of learning  6-DoF object poses for robotic grasping from RGB-D images via language instructions.   
Our model estimates  6-DoF object poses at category-level. The point cloud segmentation module helps better performance in 6-DoF pose estimation. We believe our system is significant for both robotic grasping and human-robot interaction tasks. 










\clearpage

\bibliography{./root}      
\bibliographystyle{./IEEEtranS.bst} 

\end{document}